\documentclass{article}

\usepackage{microtype}
\usepackage{graphicx}
\usepackage{subfigure}
\usepackage{booktabs} % for professional tables
\usepackage{amsmath}

\usepackage{hyperref}

\usepackage[accepted]{icml2019}

\begin{document}
\renewcommand\baselinestretch{0.9}
% \baselineskip=1.05\normalbaselineskip

\twocolumn[
\icmltitle{WeNet: Weighted Networks for Recurrent Network Architecture Search}



\icmlsetsymbol{equal}{*}

\begin{icmlauthorlist}
\icmlauthor{Zhiheng Huang}{to}
\icmlauthor{Bing Xiang}{to}
\end{icmlauthorlist}

\icmlaffiliation{to}{Amazon AWS AI}
\icmlcorrespondingauthor{Zhiheng Huang}{zhiheng@amazon.com}
\icmlcorrespondingauthor{Bing Xiang}{bxiang@amazon.com}

\icmlkeywords{Machine Learning, ICML}

\vskip 0.6in
]



\printAffiliationsAndNotice{}  % leave blank if no need to mention equal contribution

\begin{abstract}
In recent years, there has been increasing demand for automatic architecture search in deep learning. Numerous approaches have been proposed and led to state-of-the-art results in various applications, including image classification and language modeling. In this paper, we propose a novel way of architecture search by means of weighted networks (WeNet), which consist of a number of networks, with each assigned a weight. These weights are updated with back-propagation to reflect the importance of different networks. Such weighted networks bear similarity to mixture of experts. We conduct experiments on Penn Treebank and WikiText-2. We show that the proposed WeNet can find recurrent architectures which result in state-of-the-art performance. 
\end{abstract}

\section{Introduction}
In the past several years, there has been groundbreaking progress in various applications, including speech recognition \cite{hinton2012,dahl2012} and image classification \cite{lecun1998,krizhevsky2012}. This progress is primarily due to advances in deep learning, e.g. the recurrent network \cite{hochreiter1997} and convolutional network \cite{lecun1998}. Numerous research works have built upon these advances, developing new network architectures, such as VGG Network \cite{simonyan2015} and ResNet\cite{he2016}. These network architectures introduce new structures which can help to boost the system accuracy. 

On the other hand, as opposed to manually designing networks, automatic network architecture search has recently been drawing more and more attention \cite{zoph2016,pham2018,liu2018,luo2018}. These automatically discovered architectures have achieved state-of-the-art performance on image classification and language model tasks. In addition, the architecture search time has been improved from  more than 1000 GPU days to several GPU hours. In terms of search types, the architecture search can be reinforcement learning (RL) based \cite{zoph2016,zoph2017,cai2017,baker2017},  evolutionary algorithm (EA) based \cite{xie2017,miikkulainen2017,real2017,liu2018,real2018}, or the recently introduced gradient descent based \cite{liu2018,luo2018} etc. In RL-based methods, a sequence of actions are searched to build optimal networks which would be rewarded by the accuracy improvement on the development dataset. In EA-based method, search is performed through mutations and re-combinations of architectural components to get better performance. While RL and EA based methods are operated in discrete space, the gradient descent based methods can be applied to continuous space for architecture search.

In this paper, we propose a novel gradient descent based method for recurrent architecture search. We conduct experiments on a language modeling task to demonstrate the effectiveness and efficiency of the proposed method. Our contribution in this paper can be summarized as the following:
\begin{itemize}
\item We introduce weighted networks (WeNets), a specific instance of mixture of experts. We define WeNets as a number of networks which connect to the same input and output layer. Each network not only has its model parameters, but also has a weight indicating how important the network is when trained with other networks. Both the model parameters and model weights are updated by the stochastic gradient descent (SGD) during model training.
\item  We propose a simple and effective algorithm for recurrent architecture search with WeNets.  The algorithm takes a collection of randomly generated network architectures, i.e. WeNets, and performs efficient architecture search. Similar to a mini-batch in SGD, we introduce \textit{network batch size} which specifies the number of networks are processed at the same time during the network architecture search.
\item We show that an architecture found by WeNets achieves state-of-the-art results on the Penn Treebank language dataset. In addition, we demonstrate that the discovered recurrent architecture can be readily used for different datasets, for example, WikiText-2 dataset.
\end{itemize}

The remainder of the paper is organized as follows. Section 2 describes the related work. Sections 3 explains the architecture search via weight networks. We report the experimental results at Section 4 and draw the conclusion at Section 5. 

\section{Related Work}

The most relevant paper to our work is DARTS \cite{liu2018}, which is the first paper to use gradient descent for architecture search. However, there are four major differences between DARTS and our work: 1) DARTS is based on the continuous relaxation of the architecture representation, which allows efficient search of possible architectures using gradient descent. In particular, a softmax function is applied over all possible operations (such as activation functions) to generate the weighted operation results. In our method, we apply a softmax function over a collection of networks to generate weighted results. In other words, DARTS considers the entire search space during architecture search, while our method restrains the search space to a specified collection of networks. 2) DARTS utilizes parameter sharing \cite{pham2018} to make the search possible, as one cannot fit all the possible network architectures (normally more than billions) to memory. On the other hand, our algorithm does not have to share parameters in architecture search. Parameter sharing \cite{pham2018} is a useful technique to make the search more efficient in terms of memory usage, but updating on shared parameters may prevent the search from heading to the correct direction. In our experiments, the search without parameter sharing leads to better models in terms of accuracy on development datasets.  3) DARTS needs both training and validation datasets to update the model and architecture parameters, while our method needs the training dataset only. 4) DARTS requires alternate parameter updating on model parameters and architecture parameters, while our method updates them simultaneously and therefore simplifies the architecture search. Our paper is also related to \cite{luo2018} as both perform architecture in continuous space. However, \cite{luo2018} utilized the encoder, performance predictor and decoder for architecture search which is a totally different framework.

In addition, this paper is related to the literature of mixtures of experts. Since mixtures of experts were introduced more than two decades ago \cite{jacobs1991,jordan1994}, it has been applied to different types of experts including SVMs, Gaussian Processes and deep neural networks. Recently, \cite{shazeer2017} has proposed sparsely-gated mixture-of-experts layers, consisting of up to thousands of feed-forward sub-networks. A trainable gating network determines a sparse combination of these experts to use for each example. There are major differences between \cite{shazeer2017} and our work. 1) We do not use gating to infer the importance of experts; instead, we use linear weights to represent the importance of experts. 2) The experts in \cite{shazeer2017}  are feed-forward networks with identical architectures, while our work considers different network architectures. 3) The gating in \cite{shazeer2017} is part of the model and is used at inference time to determine which combination of the experts are used. Our work follows the deep network architecture search literature and uses linear weights to guide the search for an efficient network architecture. At the end of the search, a single expert network is picked and the weights are no longer used in inference.

\section{Architecture Search via Weighted Networks}

\subsection{Network Search Space} \label{sec:searchSpace}

\begin{figure}[!hbt]
    \centering
    \includegraphics[scale=0.8]{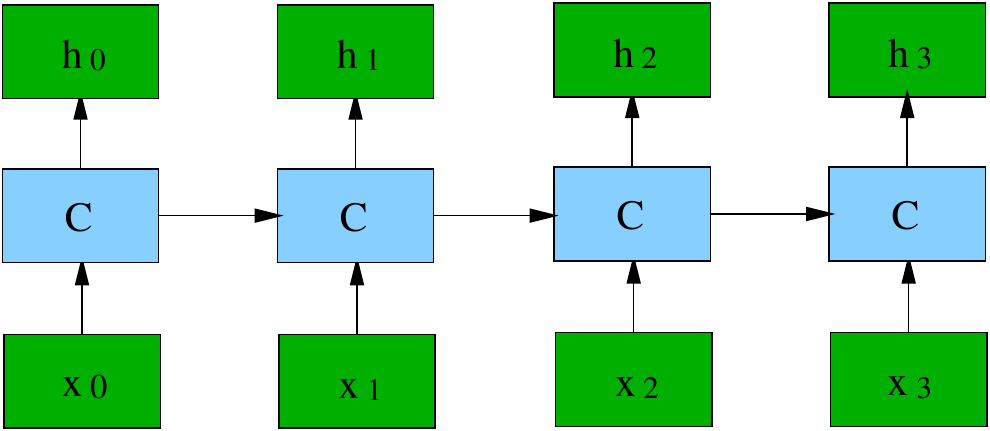}
    \caption{Recurrent network unrolled.}
    \label{fig:recurrent-unroll}
\end{figure}

Recurrent networks \cite{hochreiter1997} have been very successful in modeling sequential data, for example in language modeling \cite{bengio2003,mikolov2010,zaremba2014}. Figure \ref{fig:recurrent-unroll} shows the recurrent networks with the networks unrolled for four time steps. This chain-like nature reveals that recurrent neural networks are intimately related to sequences and lists. It is the natural choice of neural network architecture for such data. 

The core part of recurrent networks is the recurrent cell, as denoted $C$ in Figure \ref{fig:recurrent-unroll}. For each time step $t$, the cell take $x_t$ and previous hidden state $h_{t-1}$ as input and produces the output $h_t$. Some widely used cells include RNN, LSTM \cite{hochreiter1997} and GRU \cite{cho2014}. The goal of automatic machine learning (AutoML) is to automatically find a useful structure so the long-distance dependency can be modeled. 

\begin{figure}[!hbt]
    \centering
    \includegraphics[scale=0.4]{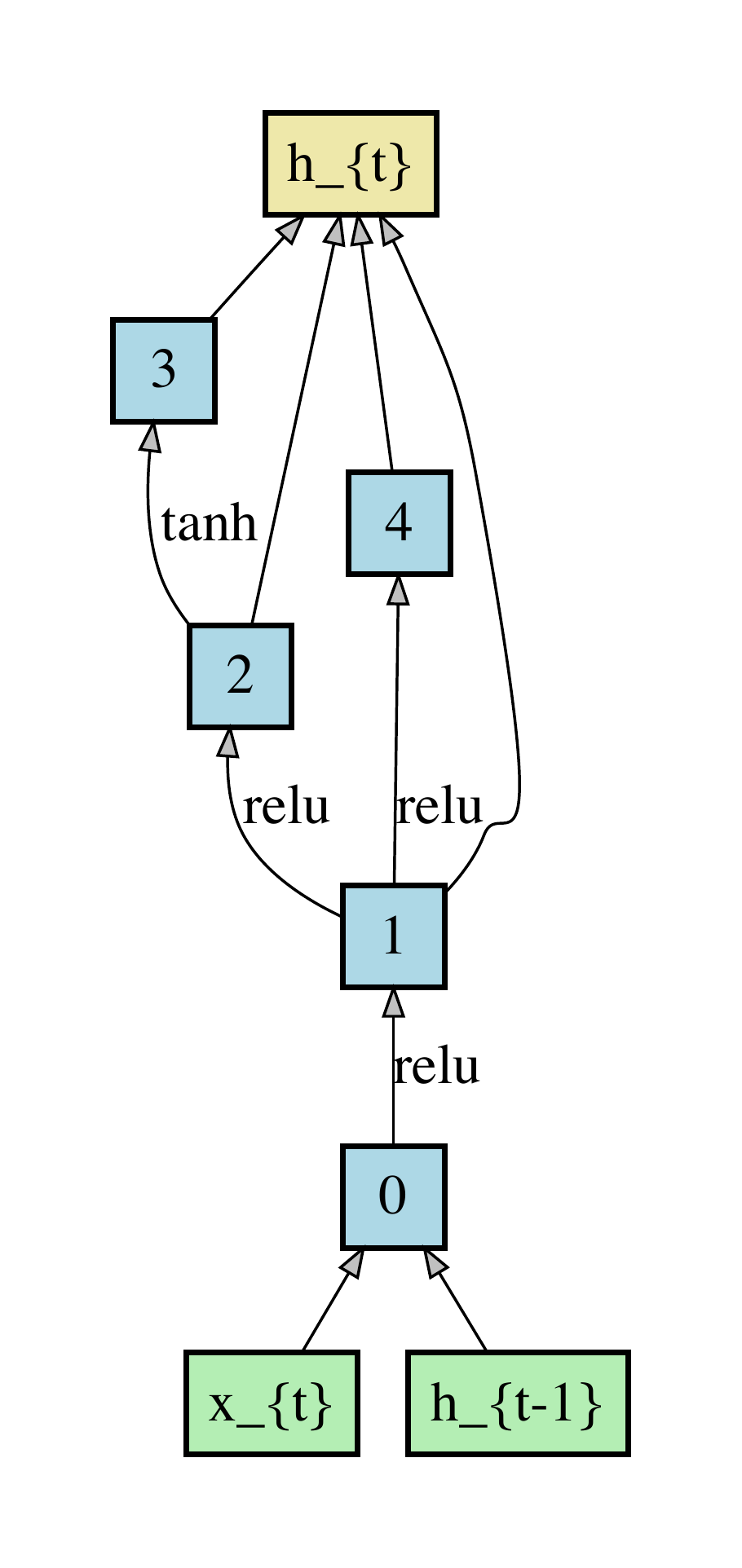}
    \caption{An example network for recurrent network architecture search.}
    \label{fig:candidateNets}
\end{figure}

Following \cite{zoph2017,real2018,liu2017a,liu2017b,liu2018}, a cell is a directed acyclic graph consisting of an ordered sequence of $L$ nodes. Figure \ref{fig:candidateNets} shows an example recurrent cell with $L=5$. 

We assume the cell has two input
nodes and a single output node. These are defined as the input at
the current step and the state from the previous step. In particular, the very first intermediate node $0$ at time step $t$ is obtained by the following transformations
\begin{eqnarray*}
c &=& \sigma(W_x x_t), \\
h &=& tanh (W_h h_{t-1}), \\
s_0 &=& h_{t-1} + c (h-h_{t-1}), \\
\end{eqnarray*}
where $\sigma$ is the logistic sigmoid function. $x_t$ is the input at time step $t$. $h_{t-1}$ is the previous hidden state. All states $c$, $h$, $h_{t-1}$, $h_t$ have the same dimensionality.  $W_x$ and $W_h$ are parameters to be learned.

The task of recurrent cell discovery is to find the ancestor node and activation function for nodes $i$, $i=1, \ldots, L-1$.  The ancestor node of node $i$ is one of the nodes which appears before $i$, that is, in the range of $0, \ldots, i-1$.  The choice of  activation functions follows \cite{zoph2016,pham2018,liu2018} and it includes tanh, relu, sigmoid, and identity mapping respectively. The state of node $i$, $s_i$, is thus computed as 

\begin{equation}
s_i = o_i(W_{ji}s_j),
\end{equation}
where $j$ is the ancestor node of $i$, $s_j$ is the state of node $j$, $o_i$ is the one of the activation functions (sigmoid, tanh, relu, etc.), and $W_{ji}$ is the model parameter to be learned. For example, the $4$-th node in Figure \ref{fig:candidateNets} is computed as follows:

 \begin{equation}
s_4 = relu(W_{14}s_1).
\end{equation}

Previous work \cite{pham2018} proposed the parameter sharing in network structure discovery. That means network parameters such as $W_{14}$ and $W_{12}$ may be shared. We compared the results between parameters sharing and non-parameters sharing. We found that non-parameter sharing results in superior results and thus we report all results in this paper without parameter sharing. We hypothesize that in the parameter sharing setting, the same model parameters are updated for different nodes and operators, thus resulting \textit{parameter updating collision}. On the other hand, the non-parameter sharing avoids the parameter updating collision and it can more accurately assess the importance of candidate networks in architecture search.

The output of the cell is obtained by averaging all the intermediate nodes:
\begin{equation}
h_t = \frac{1}{L-1} \sum_{i \in \{1, \ldots, L-1\}} s_i
\end{equation}

We note that the search space defined above does not cover the LSTM cell. It would be interesting to re-define the search space to have LSTM as one instance of the search. Nevertheless, we follow this search space as used in previous work \cite{pham2018,liu2018}. We include the state of the art LSTM performance in experiments (see Section \ref{sec:experiments}).

\subsection{Weighted Networks}

\begin{figure*}[!hbt]
    \centering
    \includegraphics[scale=0.6]{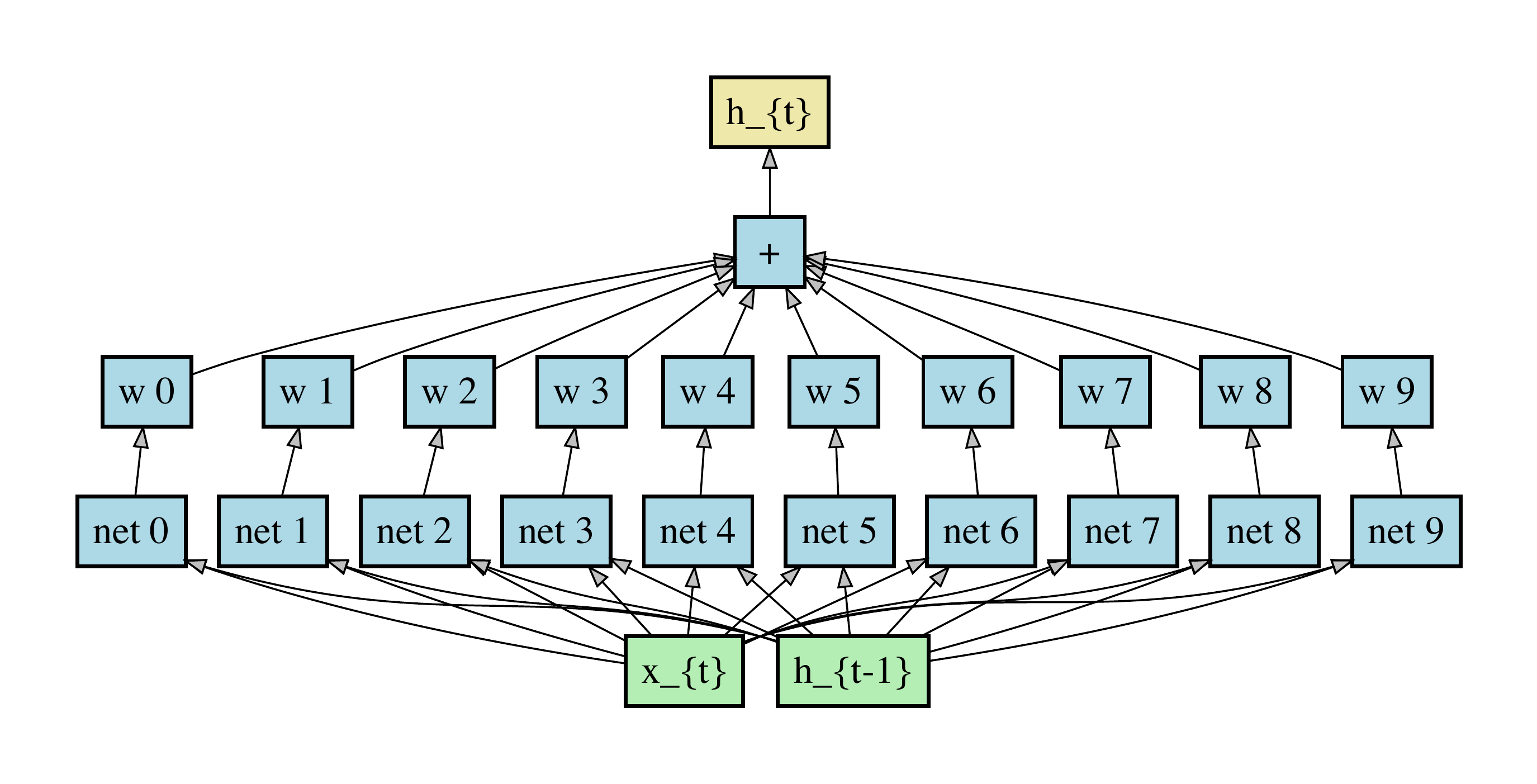}
    \caption{Weighted network structure}
    \label{fig:wenet}
\end{figure*}

The weighted networks consist of a collection of candidate networks $N_0, \ldots, N_{n-1}$, and weight parameters $w_0, \ldots, w_{n-1}$. Figure \ref{fig:wenet} shows an overview of WeNets, with 10 networks $net_i$, $i=0, 1, \ldots, 9$, considered. Each network represents a network architecture  similar to that in Figure \ref{fig:candidateNets}. As can been seen from this figure, each candidate network expects the same-sized inputs ($x_{t}$ and $h_{t-1}$) and produces the same-sized outputs ($h_{t}$). The outputs from all candidate networks are averaged to generate the final output $h_{t}$.

Let's denote $N_i(x)$ the output of candidate network $N_i$ for a given input $x$, the weighted network can be written as follows. 
\begin{equation}
y=\sum_{i=1}^{n} w_i N_i(x),
\end{equation}

where $w_i$ is the weight for network $N_i$. The number of candidate networks, $n$, is determined by the GPU memory available. In our experiments, we have up to one hundred candidate networks which can be fit to one GPU memory. Note that $w_i$ are normalized with softmax and they can be interpreted as the importance of networks.

\subsection{Architecture Search Algorithm}

The architecture search algorithm consists of two steps. The first is to randomly generate a collection of networks and the second is to search over these networks to find optimal ones.

\subsubsection{Random Networks Generation}

Algorithm \ref{algo:randomGen} shows how to randomly generate $T$ networks. The algorithm generates an empty list $res$ at line 2. It then goes to the procedure to generate each candidate net at line 5 to 8. It stops when the networks generated reach the specified number $T$. 

For each network generation, we first randomly sample a previous node to be the ancestor node to connect to. For example, if $l==2$, we randomly select a \textit{node} from node list of \{0, 1\} which were previously generated. In addition, we randomly sample an \textit{op} from the list of tanh, relu, sigmoid and identity. We insert the pair of \textit{node} and \textit{op} to the candidate network. After pairs of node and op are generated for all levels $L$, we have a complete network architecture which may be similar to the following:

\textit{[('relu', 0), ('relu', 1), ('tanh', 2), ('relu', 1)]}

We then average all of these nodes ($i = 1, \ldots, 4$) to produce the final result (see Figure \ref{fig:candidateNets}).

\begin{algorithm}[!hbt]
\begin{algorithmic}[1]
\STATE {\bfseries Input:} Total networks to generate: $T$, Recurrent network levels: $L$
\STATE $res=[]$
\WHILE{$len(res) < T$} 
\STATE Create an empty net $C =[]$
\FOR{level $l$ in L}
  \STATE $node = randInt(0, l)$
  \STATE $op = randomSampling$ 
   \STATE \ \ \ \ \ \ \ \ $\{tanh, relu, sigmoid, identity\}$
  \STATE $C.append((node, op))$
\ENDFOR
\STATE res.append(C)
\ENDWHILE
\STATE return res
\end{algorithmic}
%\end{small}
\caption{Random Network Generalization}\label{algo:randomGen}
\end{algorithm}

\subsubsection{Search Algorithm}

Once we have a collection of randomly generated networks, we can search over them with Algorithm \ref{algo:trainProcedure}. The algorithm expects the following hyper-parameters: total networks to search $T$, network batch size $B$ and network seeding size $K$. We first randomly sample $T$ candidate networks as shown in Algorithm \ref{algo:randomGen} and initialize the seed networks to be empty. We go to the loop starting with line 4 to process all networks. In particular, we take $B$ (network batch) networks from $pool$ (line 5) and combine the previous seeds to be the new candidates (line 7). We train the candidate networks jointly by the WeNet structure (Figure \ref{fig:wenet}) on training data. The duration of training is specific to applications. In language modeling, we find that one or two epochs are enough to discover good network architectures (See Section \ref{sec:experiments} for more details). After the training at line 8, out of the candidate networks, we select $K$ networks which have the maximum weights to form the new $seed$ networks. We keep running the algorithm until we process all $T$ networks. We return best network from $seed$ which has the maximum network weight.

\begin{algorithm}[!hbt]
\begin{algorithmic}[1]
\STATE {\bfseries Input:} total networks to search: $T$, network batch size: $B$, network seeding size: $K$
\STATE Randomly generates $T$ candidate networks denoted as $pool$
\STATE Initiate seed networks $seed = \{\}$
\WHILE{$pool$ is not empty}
\STATE Choose $B$ networks denoted as $candidates$ from $pool$ 
\STATE $pool = pool - candidates$
\STATE $candidates = candidates \cup seed$
\STATE Train the weighted networks of $candidates$ on training data
\STATE Update $seed$ to contain the $K$ networks which have the maximum $K$ net weights $w$
\ENDWHILE
\STATE Return best network from $seed$
\end{algorithmic}
%\end{small}
\caption{Architecture Search Procedure}\label{algo:trainProcedure}
\end{algorithm}

\section{Experiments} \label{sec:experiments}

Neural language modeling \cite{bengio2003,mikolov2010,zaremba2014} has been a fundamental task in natural language processing and speech recognition. This task has been widely used to test recurrent neural networks. We use Penn Treebank (PTB) language model dataset to test the network architecture search algorithm proposed in this paper. 

The experiments consist of the following: 1) The recurrent architecture search. We report the setup for architecture search by WeNet. 2)  The architecture evaluation. We use the architecture found in 1) to train a language model from scratch and report the performance on the test data set. 3) We investigate the transfer-ability of the found architecture on PTB by evaluating them on WikiText-2 (WT2). 4) We compare the network structure with the previously discovered DARTS structure.

\subsection{Recurrent Architecture Search on Penn Treebank}

We follow the network search space setup as in  \cite{zoph2016,pham2018,liu2018} (see Section \ref{sec:searchSpace}). We consider the selection of tanh, relu, sigmoid, and identity activation functions and the recurrent cell consists of $L = 8$ nodes. As in ENAS cell \cite{pham2018} and DARTS cell \cite{liu2018}, we enable batch normalization in each node to prevent gradient explosion during architecture search, and disable it during architecture evaluation. In network search, the recurrent network consists of only a single cell. That is, we do not use any repetitive patterns by vertically stacking the cells.

We run  Algorithm \ref{algo:trainProcedure} for recurrent architecture search. We set the total networks of $T$ to be $10k$, batch network size $B$ to be 100, and seeding size $K$ to be 20. For each network, the size of embedding and hidden units are set to 200. We use data batch size of 20.  We found that small data batch size is essential to stabilize the architecture search. We use BPTT length 35, and weight decay 5. We apply variational dropout \cite{gal2016} of 0.2 to word embeddings, 0.75 to the cell input, and 0.25 to all the hidden nodes. A dropout
of 0.75 is also applied to the output layer. Other training settings are identical to those in \cite{merity2017,yang2017}. We choose Adam as the optimizer during search.

\begin{figure*}[!t]
    \centering
    \includegraphics[scale=0.9]{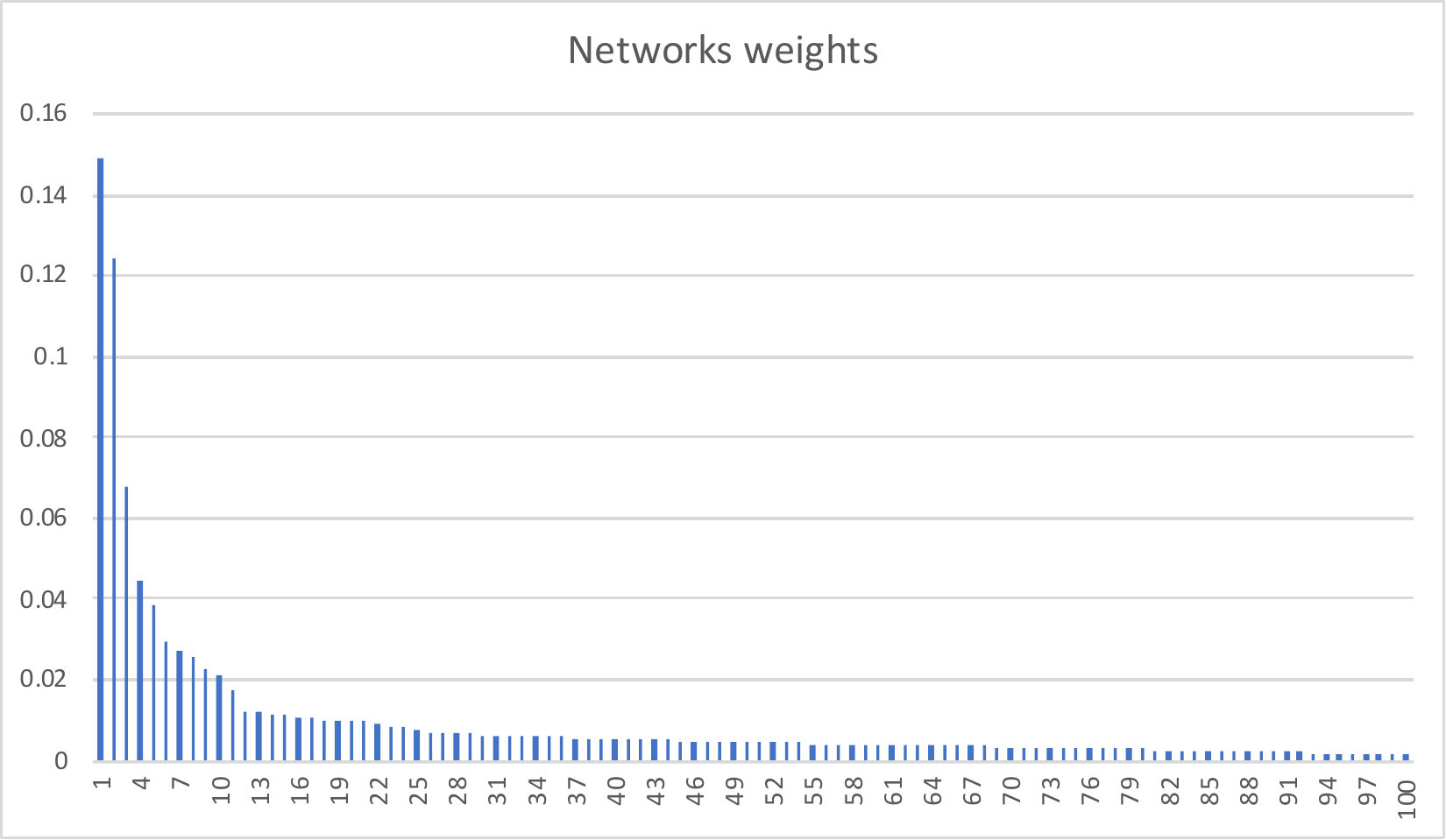}
    \caption{Network weights plot for WeNet.}
    \label{fig:net_weights}
\end{figure*}

We train WeNet for two epochs (line 8 in Algorithm \ref{algo:trainProcedure}) to select the best candidate networks according to the network weights. Figure \ref{fig:net_weights} shows an example of sorted network weights after training of two epochs. The top $20$ networks (out of $100$) are retained to continue architecture search. These weights are normalized from softmax and they can be interpreted as the importance of the networks considered. Each epoch takes about 2 minutes, and thus the whole architecture search takes about 2*2*10k/100/60 =  6 hours. Similar to the training speedup for the utilization of data batch, the WeNet search effectively introduces the concept of \textit{network batch}, and it thus provides significant search efficiency when compared to training the networks one at a time. 

\begin{figure}[!t]
    \centering
    \includegraphics[scale=0.4]{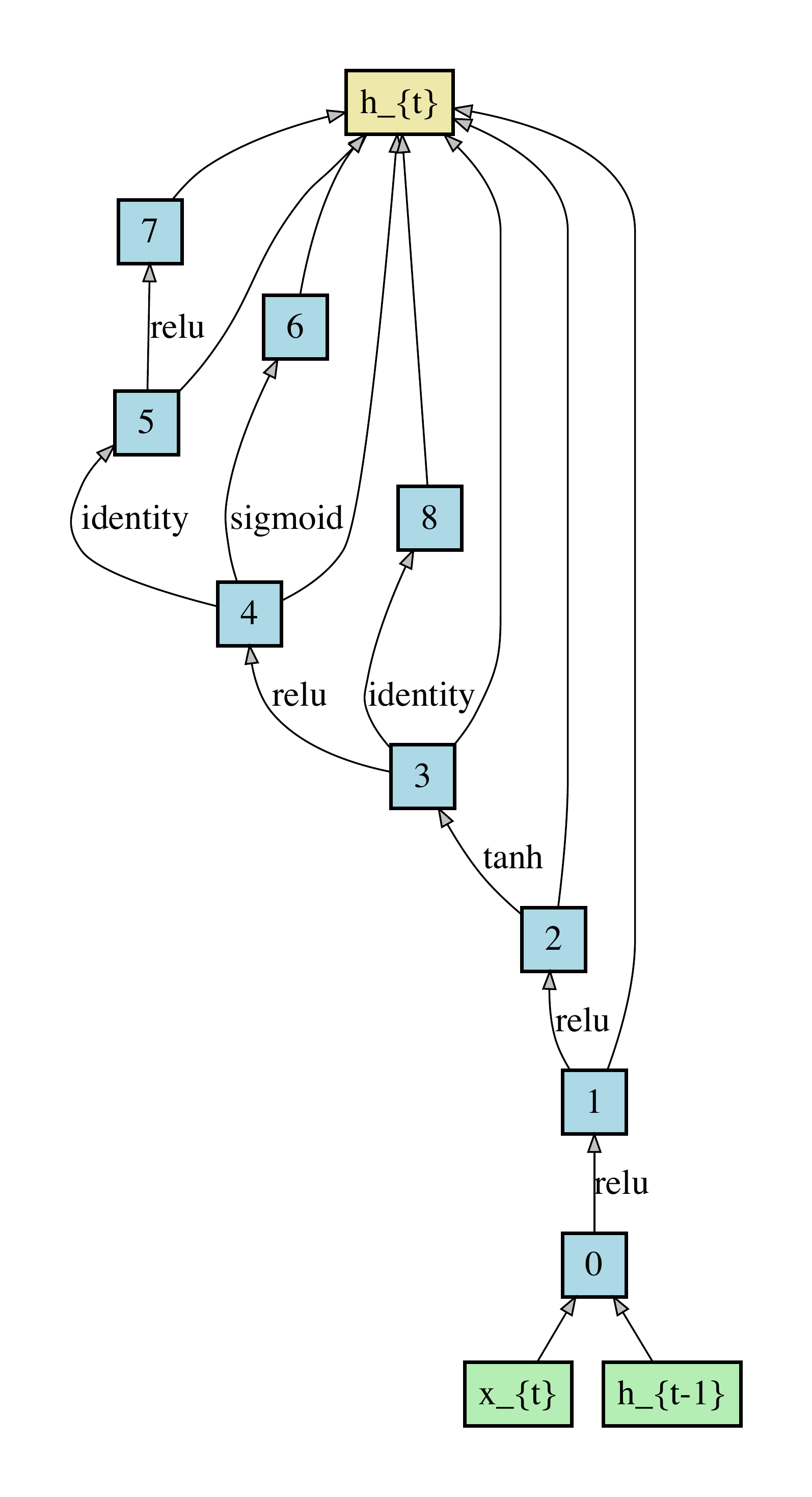}
    \caption{Best Recurrent Network Discovered by WeNet.}
    \label{fig:bestNet}
\end{figure}

Considering the random network initialization may result in fluctuation in results, we run Algorithm \ref{algo:trainProcedure} four times and report the experiments with the best found architecture, as shown in Figure \ref{fig:bestNet}.

\subsection{Recurrent Architecture Evaluation}
We follow the setup in \cite{liu2018}. A single-layer recurrent network with the discovered cell is trained for 1600 epochs, with batch size
64, averaged SGD \cite{polyak1992} (ASGD), learning rate 20, and weight
decay $8 \times10^{-7}$. To speed up, we start with SGD and trigger ASGD using the same protocol as in \cite{yang2017,merity2017}. Both the embedding and hidden unit sizes are set to 850 to
ensure our model size is comparable with other baselines. Other hyper-parameters, including dropouts,
remain exactly the same as those for architecture search. For fair comparison, we do not finetune our
model at the end of the optimization, nor do we use any additional enhancements such as dynamic
evaluation \cite{krause2017} or continuous cache \cite{grave2016}. The training takes 5
days on a single Tesla V100 GPU with our implementation. Our code is implemented on top of DARTS and thus we can facilitate fair comparison without implementation discrepancies.

Table \ref{comp_ptb} presents the results for recurrent architectures on PTB. As reported in \cite{liu2018}, the random architectures are competitive. For example, it leads to the perplexity of 61.5 on PTB test data. Nevertheless, recent work including the LSTM mixture of softmaxes, ENAS, DARTS, NAONet and WeNets are able to improve the baseline significantly. The cell discovered by ENAS, DARTS, and NAONet
achieved the test perplexity of 58.6, 56.1 and 56.0 respectively. It is worth noting that the NAONet takes much longer time for architecture search (300 GPU hours). The short version NAONet-WS results in worse perplexity ($56.6$).  The best previously reported is from \cite{yang2017} which obtained the perplexity of $56.0$ on test dataset. The DARTS has the perplexity of $56.1$ and is competitive with the previous state-of-the-art model. The architecture discovered by WeNet results in perplexity of $57.9$ when trained with 1500 epochs, which is comparable to ENAS structure but not as good as DARTS structure. However, the WeNet structure can lead to the new state-of-the-art perplexity ($54.87$) when trained with more epochs (6000), while the DARTS and ENAS nets are not able to lower the perplexity of validation dataset. In fact, we encountered gradients explosion after 2000 epochs. It is worth noting that the newly discovered architecture and the mixture of softmaxes are complimentary. They could be used jointly to improve the accuracy further. In terms of efficiency, the overall cost (4 runs in total) is within 1 GPU day, which is comparable to ENAS and DARTS and significantly faster than NAS \cite{zoph2016}. 

 \begin{table*}[!hbpt]
    \centering
    \caption{Comparison with state-of-the-art language models on Penn Treebank. Results marked with $\dagger$ were obtained in DARTs github repo. Results marked with $*$ were obtained using \cite{pham2018} public released repo.}
    \label{comp_ptb}
    \begin{tabular}{llcccccc}
    \hline
    & Architecture & valid  & test & Params (M) & Search Cost  & Search  \\ 
    &  & pplx & pplx & & (GPU days) & Method \\ \hline
    & Variational RHN \cite{zilly2016} & 67.9 & 65.4 & 23 & - & manual \\
    & LSTM \cite{merity2017} & 60.7 & 58.8 & 24 & - & manual \\
    & LSTM + skip connections \cite{melis2017} & 60.9 & 58.3 & 24 & - & manual \\
    & LSTM + 5 softmax experts \cite{yang2017} & - & 57.4 & - & - & manual \\
    & LSTM + 15 softmax experts \cite{yang2017} & 58.1 & 56.0 & 22 & - & manual \\ \hline
    & NAS \cite{zoph2016} & - & 64.0 & 25 & 1e4 CPU days & RL \\
    & ENAS \cite{pham2018} * & 68.3 & 63.1 & 24 & 0.5 & RL \\
    & ENAS \cite{pham2018} $\dagger$ & 60.8& 58.6 & 24 & 0.5 & RL \\ \hline
    & Random & 64.1 & 61.5 & 23 & - & - & \\
    & DARTS (first order) \cite{liu2018} & 62.7 & 60.5 & 23 & 0.5 & gradient-based \\
    & DARTS (second order) \cite{liu2018} & 58.8 & 56.6 & 23 & 1 & gradient-based \\
     & DARTS (second order) + 1e3 epochs \cite{liu2018} & 58.3 & 56.1 & 23 & 1 & gradient-based \\ \hline
     & NAONet \cite{luo2018} & N/A & 56.0 & 27 & 300 & gradient-based \\
     & NAONet-WS \cite{luo2018} & N/A & 56.6 & 27 & 0.4 & gradient-based \\ \hline
     & WeNet (1500 epochs) & 60.1 & 57.9 & 23 & 1 & gradient-based \\
     & WeNet (6000 epochs) & \textbf{56.8} & \textbf{54.8} & 23 & 1 & gradient-based \\
     \hline    \end{tabular}
\end{table*}

\subsection{Transfer-ability of Architectures}
In this section, we test if the recurrent architecture found on Penn Treebank can perform well on another dataset WikiText-2. We set embedding and hidden unit sizes to 700, weight decay $5\times 10^{-7}$, and hidden-node variational dropout 0.15. Other hyper-parameters remain the same as in our PTB experiments. Table \ref{comp_wt2} shows that the cell identified by WeNet transfers better than ENAS, DARTS and NAONet on WikiText-2. In particular, ENAS, DARTS and NAONet lead to the perplexities of $70.4$, $66.9$ and $67.0$ on test dataset, while WeNet results in the perplexity of $66.6$. The state-of-the-art on this dataset is from \cite{yang2017}. Lower perplexity numbers may be obtained by recurrent architecture search directly on the WikiText-2 dataset. 

 \begin{table*}[!hbpt]
    \centering
    \caption{Comparison with state-of-the-art language models on WT2. Results marked with $\dagger$ were obtained in DARTs github repo. }
    \label{comp_wt2}
    \begin{tabular}{llcccccc}
    \hline
    & Architecture & valid  & test & Params (M) & Search Cost  & Search  \\ 
    &  & pplx & pplx & & (GPU days) & Method \\ \hline
    & LSTM + augmented loss \cite{inan2017} & 91.5 & 87.0 & 28 & - & manual \\
    & LSTM + continuous cache pointer \cite{grave2016} & - & 68.9 & - & - & manual \\   
    & LSTM \cite{merity2017} & 69.1 & 66.0 & 33 & - & manual \\
    & LSTM + skip connections \cite{melis2017} & 69.1 & 65.9 & 24 & - & manual \\    
    & LSTM + 15 softmax experts \cite{yang2017} & \textbf{66.0} & \textbf{63.3} & 33 & - & manual \\ \hline    
    & ENAS \cite{pham2018} $\dagger$ & 72.4& 70.4 & 33 & 0.5 & RL \\ \hline     
    & DARTS (6000 epochs) \cite{liu2018} & 69.5 & 66.9 & 33 & 1 & gradient-based \\ \hline
    & NAONet \cite{luo2018} & N/A & 67.0 & 36 & 300 &
    gradient-based \\ \hline
     & WeNet (6000 epochs) & 69.2 & 66.6 & 33 & 1 & gradient-based \\
     \hline    \end{tabular}
\end{table*}

\subsection{Network Comparison to DARTS net}
\begin{figure}[!t]
    \centering
    \includegraphics[scale=0.4]{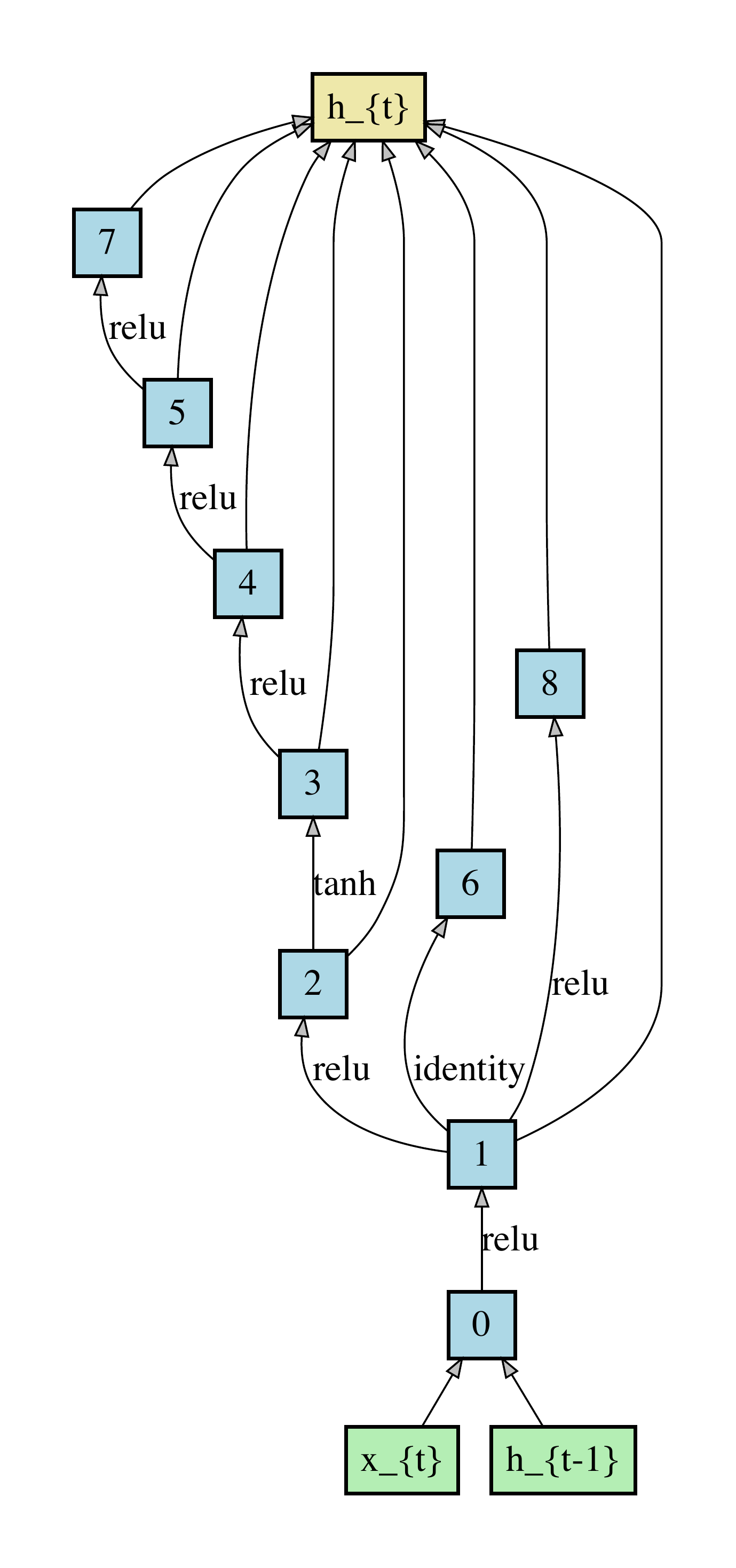}
    \caption{DARTS network}
    \label{fig:darts_v1}
\end{figure}

In this section, we compare the WeNet recurrent network structure (Figure \ref{fig:bestNet}) to DARTS (Figure \ref{fig:darts_v1}). It is interesting to see that WeNet and DARTS have a common substructure. In particular, they have exactly the same node and op pairs from level 0 to 4. We hypothesize that this subnet structure may be useful for long distance modeling. In general, DARTS and WeNet architectures demonstrate different performance behavior in training language models. DARTS outperforms WeNet structure during the first 2000 epochs of training. WeNet is able to keep improving after epoch 2000 while DARTS cannot improve further. WeNet eventually reaches the lowest perplexity at around epoch 6000. 

\section{Conclusions}
In this paper, we have introduced WeNets and proposed a simple and effective algorithm for recurrent architecture search with WeNets. We show that an architecture found by WeNets achieves state-of-the-art results on the Penn Treebank language dataset. In addition, we demonstrate that the discovered recurrent architecture can be readily used for different datasets, for example, WikiText-2 dataset.

While the proposed search algorithm is generic,  we focus on language modeling task in the paper. In the future, we would investigate the discovery of different networks including convolutional networks and sequence-to-sequence networks with WeNets. 

% \section*{Acknowledgements}

% \textbf{Do not} include acknowledgements in the initial version of
% the paper submitted for blind review.

% If a paper is accepted, the final camera-ready version can (and
% probably should) include acknowledgements. In this case, please
% place such acknowledgements in an unnumbered section at the
% end of the paper. Typically, this will include thanks to reviewers
% who gave useful comments, to colleagues who contributed to the ideas,
% and to funding agencies and corporate sponsors that provided financial

\nocite{langley00}

% \bibliography{example_paper}
\bibliography{netSearchWenet}
\bibliographystyle{icml2019}

% \appendix
% \section{Do \emph{not} have an appendix here}

% \textbf{\emph{Do not put content after the references.}}
%
% Put anything that you might normally include after the references in a separate
% supplementary file.

% We recommend that you build supplementary material in a separate document.
% If you must create one PDF and cut it up, please be careful to use a tool that
% doesn't alter the margins, and that doesn't aggressively rewrite the PDF file.
%pdftk usually works fine. 

% \textbf{Please do not use Apple's preview to cut off supplementary material.} In
% previous years it has altered margins, and created headaches at the camera-ready
% stage. 
%%%%%%%%%%%%%%%%%%%%%%%%%%%%%%%%%%%%%%%%%%%%%%%%%%%%%%%%%%%%%%%%%%%%%%%%%%%%%%%
%%%%%%%%%%%%%%%%%%%%%%%%%%%%%%%%%%%%%%%%%%%%%%%%%%%%%%%%%%%%%%%%%%%%%%%%%%%%%%%

\end{document}